\documentclass{INTERSPEECH2023}

 \interspeechcameraready

\usepackage{pgfplots}
\usepackage{colorbrewer}
\usepackage{subcaption}

\usepackage[prependcaption,textsize=scriptsize]{todonotes}

\definecolor{tableau_blue}{RGB}{31, 119, 180}
\definecolor{tableau_orange}{RGB}{255, 127, 14}
\definecolor{tableau_green}{RGB}{44, 160, 44}
\definecolor{tableau_red}{RGB}{214, 39, 40}
\definecolor{tableau_purple}{RGB}{148, 103, 189}
\definecolor{tableau_brown}{RGB}{140, 86, 75}
\definecolor{tableau_pink}{RGB}{227, 119, 194}
\definecolor{tableau_gray}{RGB}{127, 127, 127}
\definecolor{tableau_olive}{RGB}{188, 189, 34}
\definecolor{tableau_cyan}{RGB}{23, 190, 207}

\title{Acoustic Word Embeddings for Untranscribed Target Languages \\ with Continued Pretraining and Learned Pooling}

\name{Ramon Sanabria, Ondrej Klejch, Hao Tang, Sharon Goldwater}

\address{The University of Edinburgh}
\email{r.sanabria@ed.ac.uk}

\begin{document}

\maketitle
 
\begin{abstract}
Acoustic word embeddings are typically created by training a pooling function using pairs of word-like units. For unsupervised systems, these are mined using  k-nearest neighbor (KNN) search, which is slow. Recently, mean-pooled representations from a pre-trained self-supervised English model were suggested as a promising alternative, but their performance on target languages was not fully competitive. Here, we explore improvements to both approaches: we use continued pre-training to adapt the self-supervised model to the target language, and we use a multilingual phone recognizer (MPR) to mine phone n-gram pairs for training the pooling function. Evaluating on four languages, we show that both methods outperform a recent approach on word discrimination. Moreover, the MPR method is orders of magnitude faster than KNN, and is highly data efficient. We also show a small improvement from performing learned pooling on top of the continued pre-trained representations.
\end{abstract}

\section{Introduction}

Acoustic Word Embeddings (AWE) are vector representations of variable length speech segments (\textit{i.e.}, words) \cite{maas2012word, levin2013fixed}. Ideally, AWEs abstract away from non-linguistic information such as speaker gender and voice quality, so that instances of the same word cluster together in the embedding space. 
AWEs can be applied to a wide variety of search-intensive tasks such as query-by-example \cite{settle2017query} or semantic speech retrieval \cite{kamper2018semantic}, as well as in unsupervised word segmentation and clustering systems \cite{kamper2017segmental}---one step toward creating speech technology for unwritten languages. Since our eventual goal targets this task, this paper focuses on models that do not rely on transcribed speech in the target language---the unsupervised setting---and we also assume limited {\em unlabeled} target language data (up to 50 hours).

Most previous work on constructing unsupervised AWEs has approached the problem using {\em learned pooling}, where positive training pairs of similar speech segments (assumed to be the same word or n-gram) are used to learn a pooling function, based on a reconstruction \cite{kamper2019truly, peng2020correspondence,van2021comparison} or contrastive \cite{jacobs2021multilingual,robin2022speech} objective.  
Despite good AWE quality, these methods rely on identifying positive training pairs from a corpus using k-nearest-neighbors methods \cite{robin2022speech, jansen2011efficient}. Even with approximate search, such methods are computationally and memory intensive.

Recently, an alternative method for constructing AWEs was proposed  \cite{sanabria2022analyzing}, which does not rely on positive samples nor complex pooling mechanisms, but simply mean-pooling the frame-level representations from a pre-trained self-supervised model. The authors show that HuBERT \cite{hsu2021hubert} representations (pre-trained on English) 
work well for English AWEs, but less well on other languages, presumably because the model is not adapted to those target languages.

In this work, we propose solutions to the aforementioned limitations and perform experiments to directly compare the two methods individually and in combination. Specifically, for learned pooling, we show that high-quality positive pairs can be found efficiently by transcribing the target language data using a multilingual phone recognizer (MPR) trained on high-resource languages, then selecting matching phone n-grams. For average pooling of a pre-trained model, we show how to adapt English HuBERT to the target language using continued pretraining \cite{gururangan2020don,lee22i_interspeech,nowakowski2023adapting}. This entails an extra step of reconstructing HuBERT's k-means clusters, not needed for continued pretraining of wav2vec 2.0 \cite{baevski2020wav2vec}, but is worth doing since HuBERT has been shown to be more effective at word discrimination \cite{sanabria2022analyzing}.

We evaluate our AWE representations using the same-different word discrimination task (\textit{same-diff}) \cite{carlin2011rapid}
on four languages: French (which we use as a development language to select hyper-parameters and run analyses), Mandarin, German, and Xitsonga (with the latter three acting as unseen test languages from a variety of language families). 
We experiment with continued pretraining, learned pooling using a contrastive objective, and their combination.
Our experiments show that:
\begin{itemize}
    \item Using continued pretraining with 50 hours of target language data improves the performance of average-pooled HuBERT representations considerably, and most of the benefit is achieved with only 20 hours of data; 
\item For the contrastive-learning model, using MPR to identify positive pairs yields a large number of high-quality pairs, resulting in better word discrimination scores than a previous approach \cite{robin2022speech} while being orders of magnitude faster;
\item With 50h of data, continued pretraining and contrastive learning have similar performance, but contrastive learning is more data-efficient, and achieves nearly the same results with only one hour of target language data;
\item Combining both methods yields a small further improvement.
\end{itemize}

\section{Task Overview}
\label{sec:task}

In this section, we formally define our task and experimental framework. Suppose we have two utterances $x^1$ and $x^2$, both being sequences of frames.
The task (\textit{same-diff}) is to tell whether two word segments $x^1_{s:t}$ and $x^2_{s':t'}$ belong to the same word type or not.

In this paper, we focus on an approach that uses self-supervised speech representations.
We assume we have a self-supervised model $f$ to compute representations of both utterances, i.e., $z^1 = f(x^1)$ and $z^2 = f(x^2)$.
We then use a pooling function $g$ to compute a fixed-dimensional vector, also known as an AWE, to represent a word segment.
Specifically, we compute $g(z^1_{s:t})$ to represent $x^1_{s:t}$ and $g(z^2_{s':t'})$ for $x^2_{s':t'}$.
The question of whether two word segments $x^1_{s:t}$ and $x^2_{s':t'}$ are of the same word type or not becomes measuring the similarity between $g(z^1_{s:t})$ and $g(z^2_{s':t'})$ compared to other AWEs.
The cosine similarity is typically used, and the mean average precision (MAP), i.e., the area under the ROC curve when we sweep a threshold, is used as the evaluation metric.

Prior work \cite{sanabria2022analyzing} has shown strong results when $f$ is an off-the-shelf HuBERT model trained on English and $g$ is a simple averaging. In this work, we explore the setting where we have untranscribed speech of a target language to continue pre-training the self-supervised model $f$.
In addition, we also explore pooling functions with trainable parameters, such as in \cite{kamper2019truly,peng2020correspondence,robin2022speech}.
We follow \cite{jacobs2021multilingual,robin2022speech,jacobs2021acoustic} and train the pooling function $g$ with a contrastive loss.
Specifically, we use NTXent \cite{chen2020simple} which is defined as
\begin{align}
\ell_{\text{NTXent}}(c, c^+, N) &= -\log \frac{\exp(\cos(c, c^+)/\tau)}{\sum_{c^- \in N} \exp(\cos(c, c^-)/\tau)},
\label{eq:ntxent}
\end{align}
where $c$ and $c^+$ are both AWEs,
$c^+$ is a positive example for $c$, $N$ is a set of negative examples for $c$, and $\tau$ is a temperature hyperparameter.
This loss requires mining positive and negative pairs.
For example, in \cite{kamper2019truly,peng2020correspondence,robin2022speech,jansen2011efficient}, nearest neighbors are considered as positive examples, leaving others as negative examples.
Nearest neighbors are known to be slow to compute when the number of examples becomes large.
As we will see in later sections, we will explore a different approach to mining positive and negative pairs.

\section{Continued Pre-training}
\label{sec:approachcp}

In previous work, Sanabria et al.~\cite{sanabria2022analyzing} showed that for constructing AWEs on different languages, HuBERT representations (which are trained only on English) perform better than both English-trained wav2vec 2.0 \cite{baevski2020wav2vec} and multilingually-trained XLS-R \cite{conneau2020unsupervised}.
We extend their work to the setting where some 
untranscribed target language data is available.
When there is a mismatch between training and test conditions, a common approach is to continue pretraining self-supervised models on the test condition \cite{gururangan2020don,lee22i_interspeech,nowakowski2023adapting}, which motivates us to continue pretraining HuBERT on the untranscribed target language.

The task of HuBERT pretraining is masked prediction, where parts of the input are masked, and the goal is to predict the quantized speech frames of the masked parts.
To perform continued pretraining on a different language, it is not immediately obvious what training targets to use.
Similar to the original HuBERT, we run k-means on the hidden vectors from one of the HuBERT layers, and use the cluster IDs of hidden vectors as targets (described further in Section~\ref{sec:contsetup}).
We then continue to pretrain HuBERT on the target language, and finally (following~\cite{sanabria2022analyzing}), mean-pool the hidden vectors to create the AWE.

\subsection{Experimental Setting}
\label{sec:contsetup}

We evaluate our approach with the \textit{same-diff} word discrimination task \cite{carlin2011rapid} on
French, German, Mandarin, and Xistonga.
The French, German, and Mandarin sets are from Task 2 of Zerospeech 2017 \cite{dunbar2017zero}, and Xitsonga is from NCHLT \cite{barnard2014nchlt}.
Following \cite{kamperthesis}, we only use words that are at least 5 characters (or 2 characters, for Mandarin) and 0.5 seconds 
long, and we report mean average precision (MAP) scores.\footnote{Note that Xitsonga and Mandarin are tonal languages and we do not explicitly model tones or consider them during evaluation (although HuBERT may implicitly capture some tonal information). In practice, there are very few pairs of words in our data that only differ in tones, so they should not have much effect in the evaluation.} Table \ref{tab:hours} summarizes the statistics of each test set.\footnote{The list of test words, along with models and other materials used in our experiments, are available at  \url{https://github.com/ramonsanabria/awe_ssl}. }

In contrast to \cite{robin2022speech}, we avoid pretraining on the set we evaluate on, and sample 50 hours of untranscribed data
from multilingual LibriSpeech \cite{pratap2020mls} for French and German,
AIShell \cite{bu2017aishell} for Mandarin, and a separate set in NCHLT \cite{barnard2014nchlt} for Xitsonga.  We use randomly sampled utterances to increase speaker diversity, which has been shown to be important for pre-trained models \cite{sanabria2022measuring}.

We use HuBERT BASE,\footnote{\url{https://github.com/facebookresearch/fairseq/tree/main/examples/hubert}} a 12-layer Transformer, implementated in Fairseq.
We feed the untranscribed data from the target language to HuBERT, and run k-means with 500 clusters on the hidden vectors from the 10th layer of HuBERT. We use the cluster IDs of each frame as targets for continued pretraining.\footnote{We also explored using k-means centroids on MFCCs or hidden vectors of HuBERT on English, but using the k-means centroids on target languages consistently outperformed other units.}
We observe that after epoch 15, the performance stabilizes, so we do early stopping at epoch 15 for all languages. 
After continued pretraining, we average the hidden vectors from the 9th layer of HuBERT to construct the AWEs.
The hyperparameters are tuned on the French dataset, and we will evaluate how well they generalize to other languages. 

\begin{table}
  \caption{Numbers of lexical entries (word types) and word occurrences (instances) in the test set for each language.
  }
  \label{tab:hours}
  \centering
  \begin{tabular}{ l  c c  }
    \toprule
    \textbf{Language} & \# word types    & \# word instances    \\
    \midrule
    French             & 15354     &  39934     \\
    German             & 20286     &  45839               \\
    Xitsonga           & 1795      &  6384               \\
    Mandarin           & 3565      &   4132             \\
    \bottomrule
  \end{tabular}
  \vspace{-1em}
\end{table}

\subsection{Results}
 \label{sec:contres}

\begin{figure}
    \centering
    \hspace*{-0.3cm}
       \begin{tikzpicture}
    \coordinate (A) at (0,0);
    \coordinate (B) at (50,0);
    \coordinate (C) at (0,70);
    \begin{axis}[
    ybar,            
    height=3cm,       
    width=0.52\textwidth,            
    ymin=0,            
    ymax=120,           
    bar width=0.4cm, 
    y label style={at={(axis description cs:0.1,0.5)},anchor=south},    
    ylabel=MAP,            
    legend style={            
        at={(0.5,1.5)},            
        anchor=north,             
        legend columns=-1,            
        fill=none,            
        legend style={/tikz/every even column/.append style={column sep=0.3cm}},            
        fill=none,            
        draw=none,            
        xtick style={draw=none},           
    },            
    xtick=data,            
    xticklabels={French,German, Mandarin, Xitsonga},  
    enlarge x limits={abs=1.0cm}, 
    ]
        \addplot[fill=tableau_blue, nodes near coords, node near coords style={font=\tiny, anchor=south}] coordinates{(1,61.4) (2,47.9) (3,67.3) (4,85.4)};
        \addplot[fill=tableau_red, nodes near coords, node near coords style={font=\tiny, anchor=south}] coordinates {(1,40.6) (2,32.4) (3,58.5) (4,48.5)};
        \addplot[fill=tableau_brown, nodes near coords, node near coords style={font=\tiny, anchor=south, }]coordinates {(1,16.5) (2,12.6) (3,49.5) (4,50.6)};  
        \legend{HuBERT CP, HuBERT EN, HuBERT LANG}
    \end{axis}
    \end{tikzpicture}
    
    \caption{Word discrimination performance on the Zerospeech 2017 Task 2 sets for French, German, Mandarin, and Xistonga, using mean-pooled HuBERT representations from the the HuBERT BASE model. The baseline model is pre-trained on English (HuBERT EN). Our method uses 
    continued pretraining on 50 hours of speech from the target language (HuBERT CP), and we also compare to a model trained from scratch in each language (HuBERT LANG). }
    \label{fig:target_languages}
    \vspace{-1em}
\end{figure}
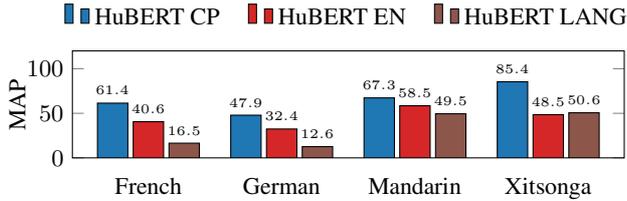

Figure \ref{fig:target_languages} shows the results for continued pretraining (HuBERT CP) on French, German, Mandarin, and Xitsonga.
We compare to the original HuBERT (HuBERT EN) and to pretraining from scratch on a particular target language (HuBERT LANG).
We observe that continued pretraining substantially outperforms others.
In addition, in three of the four cases, pretraining from scratch on the target languages underperforms
the HuBERT model pretrained on English.

These results further support the claim in \cite{sanabria2022analyzing} that by pre-training on 960 hours of English data, HuBERT \texttt{BASE} is able to learn a considerable amount of language-independent information that improves AWEs beyond what is learned from a smaller amount of target language data alone. However, it also indicates that a relatively small amount of target language data can successfully adapt the pre-trained English model and improve this type of mean-pooled AWEs. 

\section{Learned Pooling}

As we have seen, AWEs can perform well by using simple pooling functions, such as mean-pooling. To improve the system further, we study the options of learning a pooling function for constructing AWEs. As we have detailed in Section \ref{sec:task}, the goal is to learn a pooling function $g$, such that $g(z_{s:t})$ represents the segment between time $s$ and $t$ given the frame-level representation $z$ of an utterance.
Adhering to \cite{jacobs2021multilingual,jansen2011efficient,jacobs2021acoustic,algayres2020evaluating}, we focus on training the pooling function $g$ with a contrastive loss --- specifically, NTXent in \eqref{eq:ntxent}.
This loss function requires positive and negative examples that either need to be obtained from labelled data or to be mined with unsupervised approaches.
Prior unsupervised work uses nearest neighbor search for mining contrastive examples, where positive examples are taken from the near neighbors and negative examples are taken from the complement.
This approach can achieve a strong MAP, but nearest neighbor search is slow to compute and does not scale well when the data set becomes large.

We propose to use a multilingual phone recognizer (MPR) to label the untranscribed data with timings for each phone segment. Two speech segments are considered as positive pairs if they have the same phone sequence. Though the MPR system requires additional compute and data to train, we argue that the requirement is not as stringent as it seems, especially as pretrained models on high-resource languages are becoming more widely available; the use of HuBERT is one example, and the use of an external phone recognizer for unsupervised ASR in \cite{klejch2021deciphering} is another.

\subsection{Experimental Setting}

We compare two approaches to mining contrastive examples for training the pooling function: the k-nearest neighbor (KNN) approach used in previous work, and the proposed MPR approach.
For the nearest neighbor search, we first collect a set of random speech segments, ranging from 80 to 310 ms and being at least 80 ms apart.
We represent each speech segment with mean-pooled HuBERT representations, and build an approximate nearest neighbor graph \cite{jansen2011efficient,thual2018k} using dot-product as distance metric with FAISS \cite{johnson2019billion}.

The MPR system is a hybrid model based on time-delayed neural network \cite{waibel1989phoneme} trained with lattice-free maximum mutual information \cite{povey2016purely}.
The hybrid model is trained on English, Spanish, Russian, Polish, Portuguese, Bulgarian, Czech, Hausa, Swedish, and Ukrainian from GlobalPhone \cite{schultz2002globalphone}---so it has not seen any of the languages that we evaluate the AWEs on.
We collect speech segments with 2 to 5 phones. 
We define positive examples as the segments that have the same phone sequences, and negative examples as the complement. We sample a maximum of 300 n-gram instances for every n-gram type due to hard-drive limitation.

As opposed to wav2vec2.0 used in \cite{robin2022speech}, we use HuBERT throughout the experiments due to its superior performance~\cite{sanabria2022analyzing}.
We use the same network architecture as in \cite{robin2022speech} to implement the pooling function.
The network consists of a LayerNorm \cite{ba2016layer}, a 1D convolution, and a transformer layer with 4 attention heads (including position embeddings) with a learning rate of 0.0001. The model finally max pools the frame-level representations to create the AWE.
We train the pooling function (while fixing HuBERT) with a batch size of 150 for 5 epochs with a maximum 1000 iterations for each epoch.

\subsection{Results and analysis}
\label{sec:learnbased}

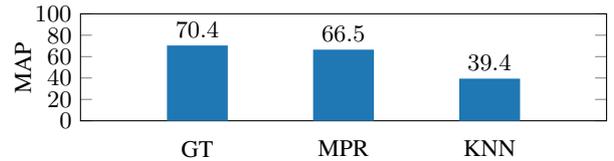
\begin{figure}
\centering
\begin{tikzpicture}
\begin{axis}[
height=3cm, 
width=0.5\textwidth,
ybar,
bar width=0.8cm,
ylabel={MAP},
y label style={at={(axis description cs:0.1,0.5)},anchor=south},
ymin=0,
ymax=100,          
symbolic x coords={GT, MPR, KNN},
xtick=data,
xlabel style={yshift=-1.5em},
xticklabel style={align=center},
xtick style={draw=none}, 
nodes near coords,
nodes near coords align={vertical},
enlarge x limits={0.4}, 
]
\addplot[fill=tableau_blue, draw=none] coordinates {(GT, 70.4) (MPR, 66.5) (KNN,39.4)};
\end{axis}
\end{tikzpicture}
    \caption{Word discrimination performance on French, our development language, with a learned pooling operation using different methods for positive pair mining. All representations use HuBERT English frame-level representations from layer 7 as input. Ground Truth (GT) and Multilingual Phone Recognizer (MPR) use 2 to 5 gram pairs. 
    \label{fig:comparission_pairs}}
\end{figure}

Results for the MPR approach and the KNN baseline are shown in Figure \ref{fig:comparission_pairs}, along with
an oracle approach that uses ground truth n-grams from forced alignment (GT).
We evaluate the approaches on 5 hours of untrascribed French, with one million positive pairs.
We find that our MPR approach performs nearly as well as using using ground truth n-grams, despite no training on the target language, and it works considerably better than the nearest neighbors approach. Moreover, for this 5h dataset, the training pairs took only five minutes to extract using MPR, versus 12 hours for the KNN approach.

\begin{figure}
\centering
\begin{tikzpicture}
\begin{axis}[
height=3cm, 
width=0.5\textwidth,
ylabel={MAP},
y label style={at={(axis description cs:0.1,0.5)},anchor=south},
ymin=0,
symbolic x coords={1, 2, 3, 4, 5, 6, 7, 8},
xtick=data,
xlabel style={yshift=-1.5em},
xticklabel style={align=center},
xtick style={draw=none}, 
]
\addplot[color=tableau_red,mark=*,smooth] coordinates {(1, 41) (2, 58.7) (3, 67.2) (4, 66.9) (5, 71.2) (6, 68.7) (7, 72.5) (8, 70.1)};
\end{axis}

\end{tikzpicture}
\caption{Word discrimination performance of contrastive-based pooling trained on ground-truth phone ngrams from different sizes. Results are reported in our development set -- French split from Zerospeech 2018 Task 2 set. We use HuBERT English frame-level representations from layer 7.}
\label{fig:ngram}
\vspace{-1em}
\end{figure}
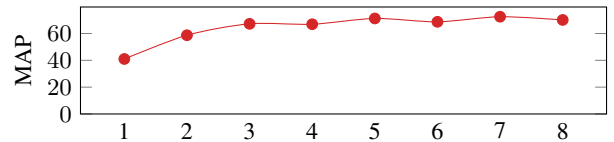

The size of the speech segments we use for mining the positive and negative examples can potentially have a significant impact on the results.
We perform a controlled experiment on the same 5 hour set with speech segments consisting of various number of forced-aligned phones to study the performance of the learned pooling function.
Results are shown in Figure \ref{fig:ngram}.
In general, larger segment sizes perform better.
Speech segments with at least 5 phones perform similarly, and we will use this setting for the rest of the experiments.

\section{Comparing and Combining Methods }
\label{sec:learnbased}

In the previous sections, we investigate configurations and show that our techniques achieve good performance. Now, we ask \textit{how do both methods compare to each other, and can they be effectively combined?} 
We also include results of the iterative nearest neighbor (IKNN) approach proposed in \cite{robin2022speech} as a baseline.
We use their implementation\footnote{\url{https://gitlab.cognitive-ml.fr/ralgayres/ssemodel}} but reduce the batch size from  250 to 150 to fit on a 12 GB GeForce GTX 1080 Ti. Instead of training with the test set as in \cite{robin2022speech}, we use a separate training set to be more comparable to our own approach. Because IKNN requires large amounts of memory, we were not able to run it on the full 50h training set, so we
report results for 20h of training data for IKNN
and also for our own best system.
We train our contrastive pooling model on 5-gram MPR positive pairs due to the results observed in Figure \ref{fig:ngram}. 

Table \ref{tab:final} presents the results on all test languages. We observe that learned pooling with pre-trained (English) HuBERT features outperforms CP the four languages, and that a small further improvement is obtained by combining the two approaches. Results are almost as good with only 20h of data, and considerably outperform the comparison approach.

\begin{table}
  \caption{
  Word discrimination results (MAP, \%) for the development language (French) and the test languages (DE=German; TS=Xitsonga; ZH=Mandarin). We compare continued pretraining  (CP); learned pooling using a contrastive objective on positive pairs from a multilingual phone recognizer (LP); the combination of both (CP + LP); and the method from~\cite{robin2022speech} (IKNN). Training uses 50h of data unless otherwise specified. 
  }

  \label{tab:final}
  \centering
  \begin{tabular}{ l c c c c }
    \toprule
    \textbf{}           & FR             & DE          &   ZH      & TS  \\
    \midrule
    CP                  & 61.4           &  47.9       & 67.3       & 85.4    \\
    LP                  & 66.6           &  70.2       & 76.1       & 89.8     \\
    CP + LP  & 69.5             &  \textbf{74.5 }      & \textbf{80.6 }      & \textbf{92.9}       \\ \midrule
    IKNN (20h)          & 41.1           &  40.0       & 52.9       & 57.8    \\
    CP + LP (20h)   & \textbf{69.6}           &  72.1       & 79.3       & 87.8        \\

    \bottomrule
  \end{tabular}
  
\end{table}

Since we are interested in a low-resource setting, we finally explore the data efficiency of each method, by reducing the amount of training data used.
We create subsets of different sizes by randomly sampling utterances from the 50-hour dataset. We use the same data to train all components (CP, K-means, and the learned pooling). 
Since longer n-gram pairs may be limited for some of the smaller data sizes, we use all 2-5 grams in all settings.

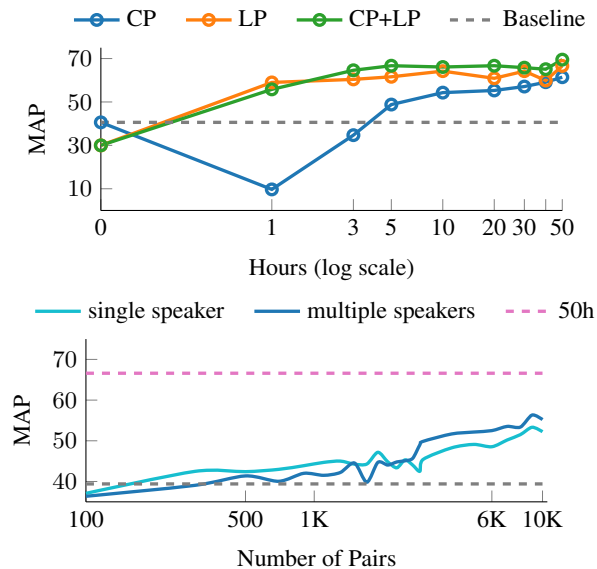
\begin{figure}
\vspace{-0.5mm}

\centering
\begin{tikzpicture}
\begin{semilogxaxis}[    
width=0.45\textwidth,    
height=0.22\textwidth, 
xmin=0.1, 
ytick={10,30,50,70}, 
xmax=50,    
ymin=0, 
ymax=75,
xlabel={Hours (log scale)}, 
ylabel={MAP},   
xtick={0.1,1,3,5,10,20,30,40,50},
xticklabels={0,1,3,5,10,20,30,,50},
legend style={at={(0.5,1.3)},anchor=north, draw=none, /tikz/every even column/.append style={column sep=0.3cm}},
y label style={at={(axis description cs:0.1,0.5)},anchor=south},
legend columns=4,
legend cell align=left,
axis x line*=bottom,
axis y line*=left
]

\addplot[color=tableau_blue,line width=1.25pt, mark=o]  coordinates {
    (0.1,40.6) (1,9.7) (3,34.7) (5,48.8) (10,54.3) (20,55.3) (30,57.1) (40,59.1) (50,61.4)
};
\addlegendentry{CP}

\addplot[color=tableau_orange,line width=1.25pt, mark=o]  coordinates {
    (0.1,30) (1,59) (3,60.4) (5,61.6) (10,64.2) (20,60.9) (30,64.2) (40,59.9) (50,66.6)
};
\addlegendentry{LP}

\addplot[color=tableau_green,line width=1.25pt, mark=o]coordinates {
     (0.1,30) (1,55.8) (3,64.6) (5,66.7) (10,66.1) (20,66.7) (30,65.8) (40,65.1) (50,69.5)
};
\addlegendentry{CP+LP}
\addplot[color=tableau_gray,line width=1.25pt, dashed,  mark=none]  coordinates {
     (0.1,40.6) (1,40.6) (3,40.6) (5,40.6) (10,40.6) (20,40.6) (30,40.6) (40,40.6) (50,40.6)
};
\addlegendentry{Baseline}

\end{semilogxaxis}
\end{tikzpicture}
\begin{tikzpicture}
\begin{semilogxaxis}[
xlabel=Number of Pairs, 
ylabel=MAP,  
xmin=100, 
xmax=10500,  
ymin=35, 
ymax=75,  
xtick={100,500,1000,6000,10000},  
xticklabels={100,500,1K,6K,10K},
ytick={10,20,...,60,70}, 
legend columns=4,
legend pos=north west,  
legend style={at={(0.5,1.3)},anchor=north, draw=none, /tikz/every even column/.append style={column sep=0.3cm}},
y label style={at={(axis description cs:0.1,0.5)},anchor=south},
width=0.45\textwidth,    
height=0.22\textwidth, 
axis x line*=bottom,
axis y line*=left
]
\addplot[tableau_cyan, smooth,line width=1.25pt] coordinates {
 (100,37.0775313) (300,42.48307672) (500,42.41207747)
  (700,42.98246323) (900,43.88024727) (1100,44.69209142)
  (1300,45.01008698) (1500,44.25935437) (1700,44.28568857)
  (1900,47.17081379) (2100,44.81645824) (2300,43.36235684)
  (2500,45.30720022) (2700,44.03111586) (2900,42.50408994)
  (3000,45.41970888) (4000,48.26873868) (5000,49.11050082)
  (6000,48.54078834) (7000,50.19255333) (8000,51.51573199)
  (9000,53.32166426) (10000,52.22393112)
};
\addplot[color=tableau_blue, smooth,line width=1.25pt] coordinates {
  (100,36.36782178) (300,39.0496066) (500,41.36783802)
  (700,40.04754011) (900,42.00589503) (1100,41.51405786)
  (1300,42.20011065) (1500,44.54206574) (1700,39.8263553)
  (1900,44.6924454) (2100,44.04126578) (2300,44.80248762)
  (2500,45.08633653) (2700,45.68545647) (2900,48.98034449)
  (3000,49.86409088) (4000,51.67938543) (5000,52.15080398)
  (6000,52.50059554) (7000,53.53413649) (8000,53.36874904)
  (9000,56.32436177) (10000,55.20072249)
};
\addplot[color=tableau_pink,line width=1.25pt, dashed,  mark=none] coordinates {
  (100,66.6) (300,66.6) (500,66.6)
  (700,66.6) (900,66.6) (1100,66.6)
  (1300,66.6) (1500,66.6) (1700,66.6)
  (1900,66.6) (2100,66.6) (2300,66.6)
  (2500,66.6) (2700,66.6) (2900,66.6)
  (3000,66.6) (4000,66.6) (5000,66.6)
  (6000,66.6) (7000,66.6) (8000,66.6)
  (9000,66.6) (10000,66.6)
};
\addplot[color=tableau_gray,line width=1.25pt, dashed,  mark=none] coordinates {
  (100,39.4) (300,39.4) (500,39.4)
  (700,39.4) (900,39.4) (1100,39.4)
  (1300,39.4) (1500,39.4) (1700,39.4)
  (1900,39.4) (2100,39.4) (2300,39.4)
  (2500,39.4) (2700,39.4) (2900,39.4)
  (3000,39.4) (4000,39.4) (5000,39.4)
  (6000,39.4) (7000,39.4) (8000,39.4)
  (9000,39.4) (10000,39.4)
};

\legend{single speaker, multiple speakers, 50h}
\end{semilogxaxis}
\end{tikzpicture}
\vspace{-1mm}
    \caption{
    Effects of training data size. (Top) Training on 1-50h of data, using mean-pooled HuBERT with continued pretraining (CP), non-CP HuBERT with learned pooling (LP), and CP using learned pooling (CP+LP); the baseline is mean-pooled (non-CP) HuBERT. All results are on the French dataset. (Bottom) Training the LP model on 100-10k pairs (from $<10$m of data), from either a single speaker or multiple speakers;  the topline is LP trained on 50h.
    }
    \label{fig:hours}
    \vspace{-4mm}
\end{figure}

Figure \ref{fig:hours} (top) presents the results. We observe that while all models reach a similar performance when trained on 50 hours of data, learned pooling achieves nearly all of the gain with only about three hours of data, indicating far greater data efficiency. This result accords with previous work: e.g., \cite{peng2020correspondence} showed that for three learned pooling methods, performance had begun to level off by 50k training pairs (the maximum they tested). It is worth noting that until \cite{robin2022speech}, it was assumed that systems should be trained on ground truth words or word-like discovered terms, leading to far fewer pairs than the n-grams used by \cite{robin2022speech} and this paper---thus, many systems were trained on only 5k-20k pairs~\cite{peng2020correspondence,van2021comparison,kamper2019truly}. With our approach, one hour of data yields about 1M pairs, and 10 minutes yields 24k pairs.

Inspired by the results above, we further explore the data requirements of the learned pooling technique, by looking at very low data regimes and the effects of speaker diversity. We sample from 100 to 10k positive training pairs, which are either randomly sampled from the 50h multi-speaker dataset, or from a single speaker. 
Figure \ref{fig:hours} (bottom) shows the results, indicating improvements over the baseline with just a few hundred training pairs, and little difference between single-speaker and multi-speaker training. These results suggest that the pre-trained HuBERT features are already doing a good job of speaker normalization, and only a relatively simple learned pooling function is needed to improve over mean-pooling. This contrasts with earlier work that learned AWEs using MFCC input features, where training pairs from multiple speakers were needed to help overcome speaker differences \cite{kamper2019truly}. 

\section{Conclusions}

We propose two techniques to adapt English self-supervised acoustic word embedding representations to a target language with up to 50 hours of unlabeled data. We first propose to adapt English frame-level representations to a target language by continued pretraining (CP). Our results using mean-pooling show show that CP is highly effective and can outperform the original model with only 10 hours of data. Next, we show that one can achieve similar performance by training a pooling mechanism on top of the self-supervied representations, using contrastive-learning with positive phone n-gram pairs obtained by a multilingual phone recognizer. The MPR method is fast and returns a large number of high-quality pairs, leading to better word discrimination than a previous approach. It is also extremely data-efficient, requiring only a few hours of target language data to reach its best results, and outperforming the previous approach with less than 1h of data. Finally, we show that the two methods can be combined, leading to the best overall results.

\bibliographystyle{IEEEtran}
\bibliography{mybib}

\begin{thebibliography}{10}
\providecommand{\url}[1]{#1}
\csname url@samestyle\endcsname
\providecommand{\newblock}{\relax}
\providecommand{\bibinfo}[2]{#2}
\providecommand{\BIBentrySTDinterwordspacing}{\spaceskip=0pt\relax}
\providecommand{\BIBentryALTinterwordstretchfactor}{4}
\providecommand{\BIBentryALTinterwordspacing}{\spaceskip=\fontdimen2\font plus
\BIBentryALTinterwordstretchfactor\fontdimen3\font minus
  \fontdimen4\font\relax}
\providecommand{\BIBforeignlanguage}[2]{{%
\expandafter\ifx\csname l@#1\endcsname\relax
\typeout{** WARNING: IEEEtran.bst: No hyphenation pattern has been}%
\typeout{** loaded for the language `#1'. Using the pattern for}%
\typeout{** the default language instead.}%
\else
\language=\csname l@#1\endcsname
\fi
#2}}
\providecommand{\BIBdecl}{\relax}
\BIBdecl

\bibitem{maas2012word}
A.~L. Maas, S.~D. Miller, T.~M. O’neil, A.~Y. Ng, and P.~Nguyen,
  ``{W}ord-level {A}coustic {M}odeling with {C}onvolutional {V}ector
  {R}egression,'' in \emph{International Conference on Machine Learning (ICML)
  Workshop}, 2012.

\bibitem{levin2013fixed}
K.~Levin, K.~Henry, A.~Jansen, and K.~Livescu, ``{F}ixed-dimensional {A}coustic
  {E}mbeddings of {V}ariable-length {S}egments in {L}ow-resource {S}ettings,''
  in \emph{Automatic Speech Recognition and Understanding Workshop (ASRU)},
  2013.

\bibitem{settle2017query}
S.~Settle, K.~Levin, H.~Kamper, and K.~Livescu, ``{Q}uery-by-example {S}earch
  with {D}iscriminative {N}eural {A}coustic {W}ord {E}mbeddings,'' in
  \emph{Interspeech}, 2017.

\bibitem{kamper2018semantic}
H.~Kamper, G.~Shakhnarovich, and K.~Livescu, ``{S}emantic {S}peech {R}etrieval
  with a {V}isually {G}rounded {M}odel of {U}ntranscribed {S}peech,'' in
  \emph{Conference on Computer Vision and Pattern Recognition (CVPR) Workshop},
  2018.

\bibitem{kamper2017segmental}
H.~Kamper, A.~Jansen, and S.~Goldwater, ``A {S}egmental {F}ramework for
  {F}ully-unsupervised {L}arge-vocabulary {S}peech {R}ecognition,'' in
  \emph{Computer Speech and Language}, 2017.

\bibitem{kamper2019truly}
H.~Kamper, ``Truly {U}nsupervised {A}coustic {W}ord {E}mbeddings {U}sing {W}eak
  {T}op-down {C}onstraints in {E}ncoder-decoder {M}odels,'' in
  \emph{International Conference on Acoustics, Speech, and Signal Processing
  (ICASSP)}, 2019.

\bibitem{peng2020correspondence}
P.~Peng, H.~Kamper, and K.~Livescu, ``A {C}orrespondence {V}ariational
  {A}utoencoder for unsupervised {A}coustic {W}ord {E}mbeddings,'' in
  \emph{Advances in Neural Information Processing Systems (NeurIPS) Wokshop},
  2020.

\bibitem{van2021comparison}
L.~Van~Staden and H.~Kamper, ``A {C}omparison of {S}elf-supervised {S}peech
  {R}epresentations as {I}nput {F}eatures for {U}nsupervised {A}coustic {W}ord
  {E}mbeddings,'' in \emph{Spoken Language Technology Workshop (SLT)}, 2021.

\bibitem{jacobs2021multilingual}
C.~Jacobs and H.~Kamper, ``{M}ultilingual {T}ransfer of {A}coustic {W}ord
  {E}mbeddings {I}mproves when {T}raining on {L}anguages related to the
  {T}arget {Z}ero-resource {L}anguage,'' in \emph{Interspeech}, 2021.

\bibitem{robin2022speech}
R.~Algayres, A.~Nabli, B.~Sagot, and E.~Dupoux, ``{S}peech {S}equence
  {E}mbeddings using {N}earest {N}eighbors {C}ontrastive {L}earning,'' in
  \emph{Interspeech}, 2022.

\bibitem{jansen2011efficient}
A.~Jansen and B.~Van~Durme, ``{E}fficient {S}poken {T}erm {D}iscovery {U}sing
  {R}andomized {A}lgorithms,'' in \emph{Workshop on Automatic Speech
  Recognition and Understanding (ASRU)}, 2011.

\bibitem{sanabria2022analyzing}
R.~Sanabria, H.~Tang, and S.~Goldwater, ``{A}nalyzing {A}coustic {W}ord
  {E}mbeddings from {P}re-trained {S}elf-supervised {S}peech {M}odels,''
  \emph{International Conference on Acoustics, Speech, and Signal Processing
  (ICASSP)}, 2023.

\bibitem{hsu2021hubert}
W.-N. Hsu, B.~Bolte, Y.-H.~H. Tsai, K.~Lakhotia, R.~Salakhutdinov, and
  A.~Mohamed, ``Hubert: {S}elf-supervised {S}peech {R}epresentation {L}earning
  by {M}asked {P}rediction of {H}idden {U}nits,'' in \emph{Transactions on
  Audio Speech and Language Processing (TASLP)}, 2021.

\bibitem{gururangan2020don}
S.~Gururangan, A.~Marasovi{\'c}, S.~Swayamdipta, K.~Lo, I.~Beltagy, D.~Downey,
  and N.~A. Smith, ``{D}on't {S}top {P}retraining: {A}dapt {L}anguage {M}odels
  to {D}omains and {T}asks,'' \emph{Association for Computational Linguistics
  (ACL)}, 2020.

\bibitem{lee22i_interspeech}
J.-H. Lee, C.-W. Lee, J.-S. Choi, J.-H. Chang, W.~K. Seong, and J.~Lee,
  ``{CTRL: Continual Representation Learning to Transfer Information of
  Pre-trained for Wav2vec 2.0},'' in \emph{Interspeech}, 2022.

\bibitem{nowakowski2023adapting}
K.~Nowakowski, M.~Ptaszynski, K.~Murasaki, and J.~Nieuwa{\.z}ny, ``{A}dapting
  {M}ultilingual {S}peech {R}epresentation {M}odel for a new, {U}nderresourced
  {L}anguage {T}hrough {M}ultilingual {F}ine-tuning and {C}ontinued
  {P}retraining,'' in \emph{Information Processing and Management}.

\bibitem{baevski2020wav2vec}
A.~Baevski, Y.~Zhou, A.~Mohamed, and M.~Auli, ``{Wav2vec 2.0: A Framework for
  Self-supervised Learning of Speech Representations},'' \emph{Advances in
  Neural Information Processing Systems (NeurIPS)}, 2020.

\bibitem{carlin2011rapid}
M.~A. Carlin, S.~Thomas, A.~Jansen, and H.~Hermansky, ``Rapid {E}valuation of
  {S}peech {R}epresentations for {S}poken {T}erm {D}iscovery,'' in
  \emph{Interspeech}, 2011.

\bibitem{jacobs2021acoustic}
C.~Jacobs, Y.~Matusevych, and H.~Kamper, ``Acoustic {W}ord {E}mbeddings for
  {Z}ero-resource {L}anguages using {S}elf-supervised {C}ontrastive {L}earning
  and {M}ultilingual {A}daptation,'' in \emph{Spoken Language Technology
  Workshop (SLT)}, 2021.

\bibitem{chen2020simple}
T.~Chen, S.~Kornblith, M.~Norouzi, and G.~Hinton, ``{A Simple Framework for
  Contrastive Learning of Visual Representations},'' in \emph{International
  Conference on Machine Learning (ICML)}, 2020.

\bibitem{conneau2020unsupervised}
A.~Conneau, A.~Baevski, R.~Collobert, A.~Mohamed, and M.~Auli, ``{Unsupervised
  Cross-lingual Representation Learning for Speech Recognition},'' in
  \emph{Interspeech}, 2020.

\bibitem{dunbar2017zero}
E.~Dunbar, X.~N. Cao, J.~Benjumea, J.~Karadayi, M.~Bernard, L.~Besacier,
  X.~Anguera, and E.~Dupoux, ``The zero resource speech challenge 2017,'' in
  \emph{Automatic Speech Recognition and Understanding Workshop (ASRU)}, 2017.

\bibitem{barnard2014nchlt}
E.~Barnard, M.~H. Davel, C.~van Heerden, F.~De~Wet, and J.~Badenhorst, ``{T}he
  {NCHLT} {S}peech {C}orpus of the {S}outh {A}frican {L}anguages,'' in
  \emph{Workshop Spoken Language Technologies for Under-resourced Languages
  (SLTU)}, 2014.

\bibitem{kamperthesis}
H.~Kamper, ``{Unsupervised Neural and Bayesian Models for Zero-resource Speech
  Processing},'' in \emph{Thesis}, 2017.

\bibitem{pratap2020mls}
V.~Pratap, Q.~Xu, A.~Sriram, G.~Synnaeve, and R.~Collobert, ``{MLS: A
  Large-scale Multilingual Dataset for Speech Research},'' in
  \emph{Interspeech}, 2020.

\bibitem{bu2017aishell}
H.~Bu, J.~Du, X.~Na, B.~Wu, and H.~Zheng, ``{AIshell-1: An Open-source Mandarin
  Speech Corpus and a Speech Recognition Baseline},'' in \emph{Oriental
  COCOSDA}.

\bibitem{sanabria2022measuring}
R.~Sanabria, W.-N. Hsu, A.~Baevski, and M.~Auli, ``{Measuring The Impact of
  Individual Domain Factors in Self-supervised Pre-training},'' in
  \emph{International Conference on Acoustics, Speech, and Signal Processing
  (ICASSP), Workshop}, 2023.

\bibitem{algayres2020evaluating}
R.~Algayres, M.~S. Zaiem, B.~Sagot, and E.~Dupoux, ``{Evaluating The
  Reliability of Acoustic Speech Embeddings},'' \emph{Interspeech}, 2020.

\bibitem{klejch2021deciphering}
O.~Klejch, E.~Wallington, and P.~Bell, ``{Deciphering Speech: a Zero-Resource
  Approach to Cross-Lingual Transfer in ASR},'' in \emph{Interspeech}, 2022.

\bibitem{thual2018k}
A.~Thual, C.~Dancette, J.~Karadayi, J.~Benjumea, and E.~Dupoux, ``{A K-nearest
  Neighbours Approach to Unsupervised Spoken Term Discovery},'' in \emph{Spoken
  Language Technology Workshop (SLT)}, 2018.

\bibitem{johnson2019billion}
J.~Johnson, M.~Douze, and H.~J{\'e}gou, ``{Billion-scale Similarity Search with
  GPUs},'' in \emph{Transactions on Big Data}, 2019.

\bibitem{waibel1989phoneme}
A.~Waibel, T.~Hanazawa, G.~Hinton, K.~Shikano, and K.~J. Lang, ``{Phoneme
  Recognition Using Time-delay Neural Networks},'' \emph{International
  Conference on Acoustics, Speech, and Signal Processing (ICASSP)}, 1989.

\bibitem{povey2016purely}
D.~Povey, V.~Peddinti, D.~Galvez, P.~Ghahremani, V.~Manohar, X.~Na, Y.~Wang,
  and S.~Khudanpur, ``{Purely Sequence-trained Neural Networks for ASR based on
  Lattice-free MMI},'' in \emph{Interspeech}, 2016.

\bibitem{schultz2002globalphone}
T.~Schultz, ``{GlobalPhone: a Multilingual Speech and Text Database Developed
  at Karlsruhe University},'' in \emph{Conference on Spoken Language
  Processing}, 2002.

\bibitem{ba2016layer}
J.~L. Ba, J.~R. Kiros, and G.~E. Hinton, ``Layer normalization,''
  \emph{pre-print}, 2016.

\end{thebibliography}

\end{document}